\documentclass[conference]{IEEEtran}
\IEEEoverridecommandlockouts
\usepackage{cite}
\usepackage{amsmath,amssymb,amsfonts}
\usepackage{algorithmic}
\usepackage{graphicx}
\usepackage{textcomp}
\usepackage{xcolor}
\usepackage{url} 
\usepackage{amsmath}

\def\BibTeX{{\rm B\kern-.05em{\sc i\kern-.025em b}\kern-.08em
    T\kern-.1667em\lower.7ex\hbox{E}\kern-.125emX}}
\begin{document}

\title{Understanding Foundation Models: \\Are We Back in 1924?  
\thanks{This research was conducted with financial support of Science Foundation Ireland 12/RC/2289\_P2 at Insight the SFI Research Centre for Data Analytics at Dublin City University.}
}

\author{
  Alan F. Smeaton\\
  Insight SFI Centre for Data Analytics,\\ 
  Dublin City University, Ireland.\\
  alan.smeaton@dcu.ie\\
  ORCID 0000-0003-1028-8389\\
}

\maketitle

\begin{abstract}
This position paper explores the rapid development of Foundation Models (FMs) in AI and their implications for intelligence and reasoning. It examines the characteristics of FMs, including their training on vast datasets and use of embedding spaces to capture semantic relationships. The paper discusses recent advancements in FMs' reasoning abilities which we argue cannot be attributed to increased model size but to novel training techniques which yield learning phenomena like grokking. It also addresses the challenges in benchmarking FMs and compares their structure to the human brain. We argue that while FMs show promising developments in reasoning and knowledge representation, understanding their inner workings remains a significant challenge, similar to ongoing efforts in neuroscience to comprehend human brain function.
Despite having some similarities, fundamental differences between FMs and the structure of human brain warn us against making direct comparisons or expecting neuroscience to provide immediate insights into FM function.
\end{abstract}


\section{Introduction}

Foundation Models or Frontier Models (FMs) are having an  impact across society which is comparable or perhaps exceeds any other technological development in AI since the earliest days of computation. To add even more significance and importance to this, the rate of development of FMs is at a pace that is difficult for us to keep up with. This means that we should try to identify the trends rather than get bogged down in the minute details and incremental developments that new releases of FMs bring. 

It is natural for us to want to look beyond the short-term future and predict where these trends might take us, should they continue in the same directions.

In this paper we identify and examine some of those trends and weave them together to identify emerging patterns in Foundation Model development, their implications for artificial intelligence capabilities, and potential future directions in this rapidly evolving field.

\section{Foundation Models}

Here we provide a quick re-cap on the most important characteristics of Foundation Models, more recently sometimes referred to as Frontier Models, that term being used to capture the characteristics of the more advanced models being at the boundary of current performance levels.

Foundation Models are large-scale AI models trained on extensive unannotated datasets which generate statistical representations of the distributions of any 1, 2, 3 or even multi-dimensional streams of data~\cite{Zhao2023ASO}. They create an embedding space from their training data which is a parametric memory, so called because the models are just a set of weights or parameters. Where the training data is text, these embedding spaces allow words, phrases or sentences to be represented as vectors in that space and the vectors capture the semantic relationships among tokens in the training data. In this way the resulting models infer a higher level of representation of information than their original source materials. 

The mapping from the training data to the Foundation Model is done using the Transformer Architecture, introduced in~\cite{vaswani2017attention} and for the largest of the large language models (LLMs), which are a type of foundation model, the computation time for this is very large.  For example GPT-4 from OpenAI is believed to have about 13 trillion tokens as its training data which, if printed as a series of books and stacked like on a shelf, round be more than 650km in length. Its compute time is estimated at $2.15 \times 10^{25}$ FLOPs which would be more than 2.5 million years on an Apple MacBook with an M2 chip.  The training of a foundation model has no natural endpoint and LLM training is usually stopped when there are negligible performance improvements on validation sets which are a form of testing how well the model would perform if training was stopped, or when the target performance levels are reached or there are signs of overfitting, or when the model's time-to-release considerations are taken into account.

Most big tech companies now have their own Foundation Models or LLMs, some are proprietary while others are open, most have a free and a subscription base for accessing them and many have multiple versions of their models which vary in model size and many of the companies compete with each other based on model size, which is reminiscent of the way search engines competed with each other based on the sizes of their indexes in the early 2000s before ultimately deciding that index size did not matter~\cite{seymour2011history}.

There is a dark side to some companies sources of training data for their LLMs. For example in 2024 alone in January we saw the Italian DPO formally charge OpenAI with violations of GDPR, in February Sarah Silverman sued OpenAI for violation of copyright on her memoir, in March the French competition authority fined Google over breaches of copyright in their training data, in April 8 US newspapers sued OpenAI and Microsoft for copyright violations, in May Scarlett Johansson forced OpenAI to stop using a voice uncannily like hers in their chat products, and the examples go on and on.

One of the concerns that users of non-proprietary LLMs have is that when the models are fine-tuned by uploading their own documents to those models, which are not stored locally but hosted elsewhere, that their fine-tuned materials, or even the original training material, could be re-constituted. K-eidetic memorisation is the concept of how many times {\it (k)} does a fact appear in the training data for a LLM and as $k$ increases the chances of the fact appearing in the synthetic output from a LLM also increases~\cite{zhang2023counterfactual}.  If a piece of data has a high value of $k$ like the home address of the US President or the company watermark on Shutterstock images then it can appear in the output because it is learned, whereas my own home address has a low $k$-eidetic, will not be learned in the model, and will not appear.  A second way in which a fact can appear is the context around a piece of data, for example my home address.  If my home address appears many times in training data because, for example, it has been the site of a well-reported robbery or it is the address of a popular business then this large amount of context around that piece of data may allow it to be re-constituted.

\section{Foundation Models and Intelligence}

\begin{figure}[ht]
    \centering
    \includegraphics[width=0.3\textwidth]{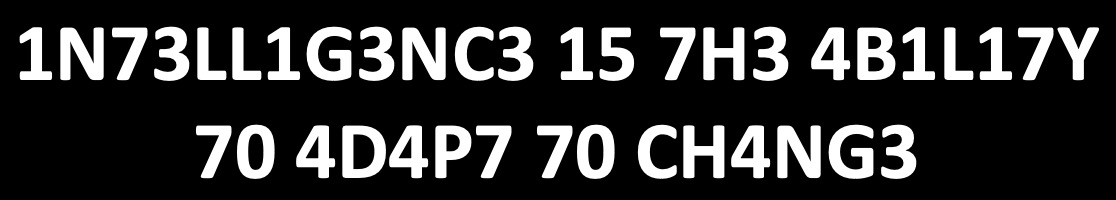}
    \caption{First fragment of text in Leet Speak.}
    \label{fig:leet1}
\end{figure}

\begin{figure}[ht]
    \centering
    \includegraphics[width=0.3\textwidth]{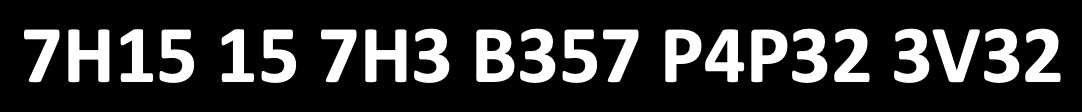}
    \caption{Second fragment of text in Leet Speak.}
    \label{fig:leet2}
\end{figure}

Consider the fragment of text shown in Figure~\ref{fig:leet1}. Can  the reader decipher it correctly? Probably yes. Now consider the fragment of text in Figure~\ref{fig:leet2}, can the reader decipher it?  Probably yes, but only because you deciphered the first one.  If you gave the text in Figure~\ref{fig:leet1} to an early version of a LLM, say GPT3, then it would correctly decipher it and  if you then gave the model the text in Figure~\ref{fig:leet2} then it would correctly decipher that one too. However, if you reversed the order and gave Figure~\ref{fig:leet2} followed by Figure~\ref{fig:leet1} the model would not decipher the text in Figure~\ref{fig:leet2} but would decipher the text in Figure~\ref{fig:leet1}.

These texts are in a form of informal language known as leet speak which originated in the 1980s and was used in online gaming and internet forums to bypass the  very simple text filters that were used then. It involves replacing letters with numbers or  characters that resemble the shapes of the original letters. For example ``E" might be replaced with ``3" and ``A" with ``4".  The leet speak version of the quote that ``intelligence is the ability to adapt to change", which is generally attributed to Stephen Hawking, appears many times on the internet\footnote{The string appears 95,800 times on the internet according to Bing}. It has thus been crawled and used as part of the training data for GPT3 so it was straightforward for GPT3 to correctly decrypt the text in Figure~\ref{fig:leet1} and when then presented with the text from Figure~\ref{fig:leet2} it it used the context of decrypting the first string, to correctly decrypt the second string. 

The string in Figure~\ref{fig:leet2}, ``7H15 15 7H3 B357 P4P32 3V32" does not appear on the internet at all according to Bing and GPT3 does not have the commonsense reasoning ability to correctly decrypt the  text from  Figure~\ref{fig:leet2} without the context of having already decrypted text from Figure~\ref{fig:leet1} (but the reader hopefully had~!). If you give the text from Figure~\ref{fig:leet2} to a more recent model like Claude-3 or GPT-4 then it correctly decrypts it   without needing the context of the text from Figure~\ref{fig:leet1}.

So what has happened in the last 18 months to allow Foundation Models to exhibit greater commonsense reasoning?
Elsewhere in AI it is well-established that larger neural networks based on larger datasets used to train them, work better in applications like speech recognition~\cite{abdel2014convolutional} and  machine translation~\cite{lu2007improving} and they out-perform conventional, symbolic and rule-based approaches to AI.
The early stage LLMs were already good at one or two-step logical inferencing, some forms of abductive reasoning, anaphora resolution and pronoun matching, and even sentiment analysis, and some other generative natural language text applications.
This is because neural networks, and Foundation Models, are a form of compression, compressing logic, reasoning and inference into their structure and weights in ways we do not understand, because we do not yet understand how a neural network operates.
So when the very recent giant Foundation Modes demonstrate logic-based processing we assume this is by dint of the volume of their training data, but how is it achieved?

GPT-4 was the first model to really show larger LLMs getting better at reasoning and that was attributed to their greater statistical basis.
In 2023
Mistral-7B~\cite{jiang2023mistral} was released by a French startup company and although only 7B parameters in size  it  beat LLaMA-1 on some inference tasks despite being only 20\% of its size. Mistral-7B runs on a laptop and is trained using grouped-query attention (GQA) and a variable size sliding  window attention (SWA) so it prunes the model size and is efficient. This was the first evidence that models which are smaller in size can not only be more efficient to use, which is a target  outcome, but counter-intuitively can also perform better at reasoning tasks.

Grokking~\cite{liu2022towards} is a phenomenon whereby large LLMs are trained and trained, and may even exhibit overfitting, and then they abruptly transition to near perfect performance on some task, and we do not know why or how this happens.  The theme of overfitting and generalisation beyond overfitting is not a topic specific to Foundation Models and has been around for several decades~\cite{karystinos2000overfitting} so it is not discussed further here.
It is accepted that models with smaller footprint sizes will need more training time and more training iterations in order to keep within their smaller footprint sizes and to converge and  stop and still perform at a similar level to larger models given the same training data.
It is believed that (over-) training of smaller models in order to stay within their size footprints yields grokking.

Even the largest Foundation Models will compress their training data since even mega-size models are much smaller than their training data.
This is achieved by learning key patterns in training data - just as humans learn language through grammar, Foundation Models codify learned patterns as combinations of weights into their parametric memory.
This can be seen as a leap from mere memorisation of training data to representing a deeper form of that training data, an equivalent of learning a grammar for a language, identifying and codifying patterns or rules in the training data, almost a form of understanding, though that would be to anthropormorphise the model which is not a good idea.
This is the learning moment that we all experience ourselves when we try to understand something complex, like how does quantum computing work and suddenly it all clicks together and we understand.  It is also  the trigger for the discussion / hype about AGI.

When FMs are used for inference, it is observed that only parts of the model are active in the computation~\cite{10.1145/3649506} and 
some of the research in the area concentrates on how to  how to reduce model size, or to prune out the unused parts of the model during inference. This is partly driven by the motivation to reduce energy and footprint sizes and goes beyond the software engineering techniques which can reduce model size like number quantization~\cite{xiao2023smoothquant}.
But when trying to understand the structure and operation of language models we do so by probing it as it is working and seeing which parts of the model are active.

Understanding a model from observations of its behaviour under different input situations is called {\it Mechanistic Interpretability}~\cite{bereska2024mechanistic} and involves identifying the patterns which have been created during training in the hidden layers and which are associated with generating different outputs.
Researchers at Anthropic  trained a model in 2024 to observe Claude's firing neurons within the model and discovered that combinations of active neurons in Claude are monosemantic~\cite{templeton2024scaling}. That means that the model has learned some patterns from its training material and has  encoded those patterns as unique combinations of weights within the model, i.e. the combination of neurons has just one semantic interpretation, it is monosemantic. To some this claim that vector spaces which have been endowed only with summation and multiplication relations and are a parametric memory of their training data, can hold semantic relations does not sit well and it is counter-intuitive that they do so. Yet the development of LLMs of considerable size has manu such conundrums, grokking being another example, and all this because we just do not understand LLMs, yet.

In the Anthropic work reported in~\cite{templeton2024scaling}, this sequence of 
\begin{align}
model~inputs & \rightarrow observe~active~neurons \nonumber \\
            & \rightarrow identify~features~among~active~neurons \nonumber \\
            & \rightarrow regenerate~original~inputs \nonumber \\ \nonumber
\end{align}  
\noindent 
involves generating another model to identify the features or characteristics of the active neurons, associate them with the model inputs and try to generate those original inputs from the active neurons.  This is the probing of a Foundation Model to see which parts are active and as we see later it is exactly the same technique as first used 100 years ago when we try to understand the human brain.
Before exploring that we will now look at the various benchmarks which are available to evaluate Foundation Model abilities to do reasoning.

\section{Benchmarking Foundation Models}

For evaluating Foundation Models’ abilities to handle different types of common-sense reasoning, several benchmarks exist as outlined in \cite{carolan2024reviewmultimodallargelanguage}, including the following:
\begin{itemize}   
\item AI2 Reasoning Challenge (ARC): This test assesses knowledge and common-sense reasoning through grade-school level multiple choice questions~\cite{clark2018think}.
\item HellaSwag: This test evaluates common-sense reasoning by requiring models to complete sentences based on everyday events, and in this way it evaluates natural language inference~\cite{zellers2019hellaswag}.
\item BoolQ: This benchmark consists of real yes/no questions from Google searches paired with Wikipedia passages. It challenges models to infer answers from context that may be implied but not stated~\cite{clark2019boolq}.
\item OpenBookQA: This question-answering dataset has been modelled after open book exams used for assessing human understanding of various subjects~\cite{mihaylov2018can} and is thus an evaluation of knowledge retrieval.
\item PIQA (Physical Interaction Question Answering): This benchmark evaluates a models’ knowledge and understanding of the physical world by presenting hypothetical scenarios with specific goals and multiple choice solutions~\cite{bisk2020piqa}.
\item Multitask Language Understanding (MMLU): This benchmark measures LLM knowledge across multiple different subject areas using multiple choice questions~\cite{hendrycks2020measuring}.
\item TruthfulQA~\cite{lin2022truthfulqa} is designed to assess the truthfulness of a model’s responses.  It achieves this by querying a model's responses on 817 questions of various styles  across a range of 38 diverse categories, intentionally constructed to challenge both comprehension and accuracy. The output generated by the model is then scrutinised for  signs of misinformation. 
\item M-HALDetect~\cite{gunjal2023detecting} serves as a dataset specifically tailored for evaluating a model’s tendency for object hallucinations. This benchmark is important for identifying instances where the model may generate outputs containing false or misleading information.
\end{itemize}

While this is a large range of benchmarks, almost all focus on some single aspect of a model's response, like truthfulness, knowledge or reasoning.  Another weakness is that the question of  polluting the test data by  the use of training data in order to achieve high scores which undermines any subsequent LLM/FM comparison.
Benchmarking to assesses overall answer quality is challenging as systems struggle to accurately measure the value of free-form responses as such comparisons cannot be  based on pairwise comparisons since there is no groundtruth against which to compare.

\begin{figure*}[ht]
    \centering
    \includegraphics[width=0.8\textwidth]{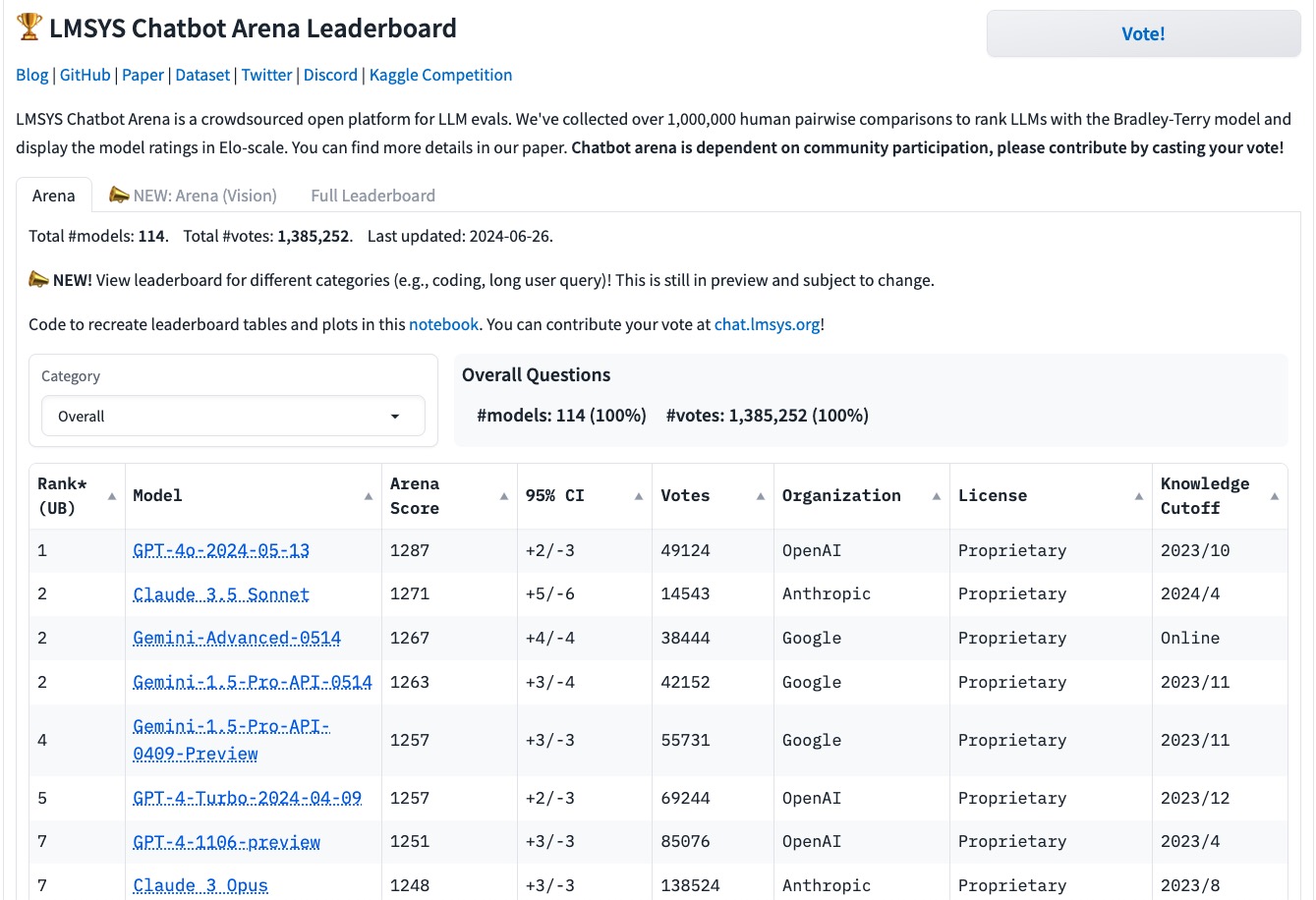}
    \caption{Screengrab from the Chatbot Arena crowd sourced evaluation platform, taken from~\cite{CBA}.}
    \label{fig:CBA}
\end{figure*}

A crowdsourced platform called Chatbot Arena~\cite{CBA} \cite{chiangchatbot} run by the Large Model Systems Organization (LMSYS Org) does address overall model quality by including  human-in-the-loop evaluation. A screenshot of Chatbot Arena is shown in Figure~\ref{fig:CBA}. It uses the Elo rating system~\cite{elo1961new} widely used in chess and other pairwise competitive games, and at the time of writing it ranks 114 LLMs (though some are just fine-tuned major LLMs) by pairing them against each other in ``battles”. 
Battles occur where a (real) user prompts the system, a random selection of two models respond, the user rates the responses and the user judgement is fed into the evaluation. As of June 2024 more than 1.3 million battles have taken place and while this is a significant amount of feedback what is evaluated by the humans-in-the-loop is really LLM popularity rather than LLM quality.

Another third party platform for evaluating LLMs, including multimodal (vision-language) models, is OpenCompass 2.0~\cite{OpenCompass}. This benchmarks more than 100 LLMs, has more than 100 datasets and can perform up to 29 core tasks on those datasets via 400,000 questions. The tasks
include information retrieval, intention recognition, sentiment analysis, summariziation, critiquing, machine translation, traditional cultural understanding, Chinese semantic understanding,  and multi-turn conversation. OpenCompass is
a toolchain with a browser interface to a collection of evaluation tools and it culminates in a set of leaderboards incorporating both open-source and proprietary benchmarks.
OpenCompass is open-source and reproducible and its evaluation leader boards are based on performance averaged across multiple datasets  using multiple evaluation metrics.

\begin{figure*}[ht]
    \centering
    \includegraphics[width=0.9\textwidth]{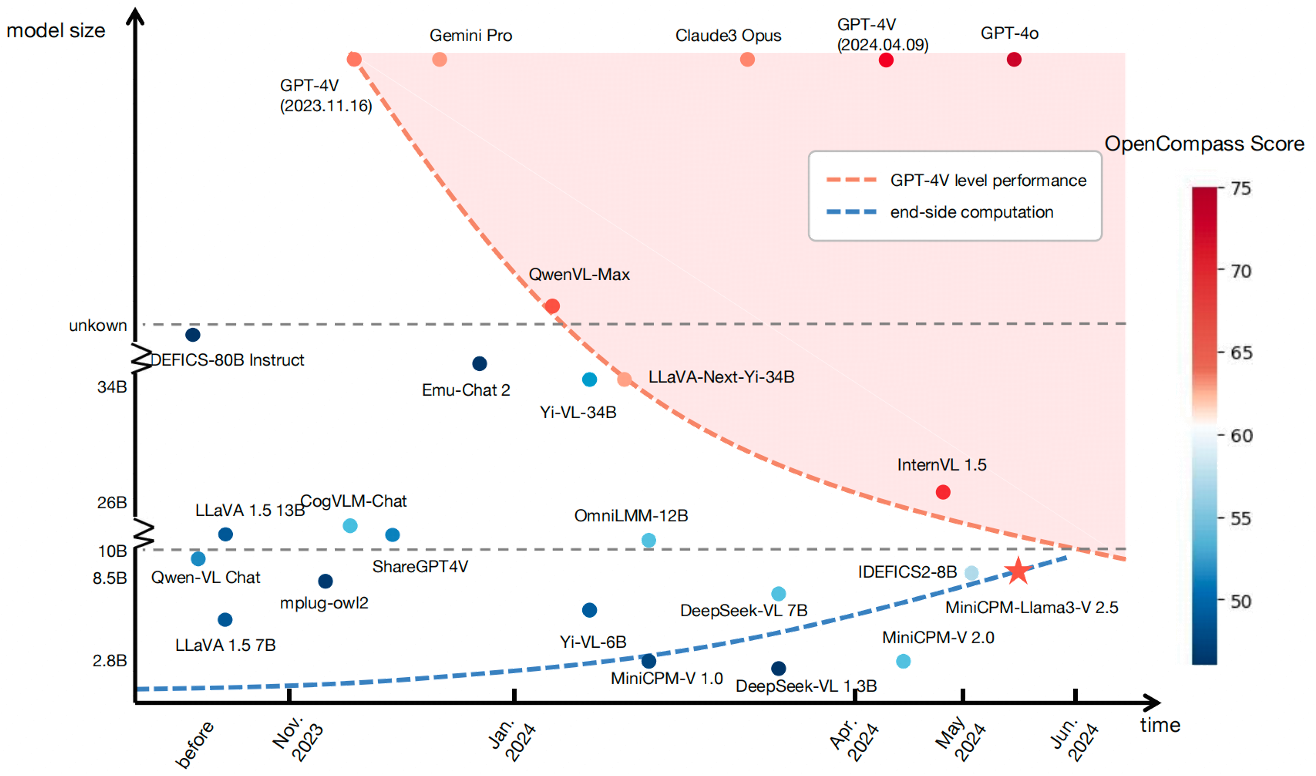}
    \caption{OpenCompass performance of a range of recently released LLMs of differing model sizes measured against the performance of GPT-4V, taken from~\cite{yao2024minicpm}.}
    \label{fig:OpenC}
\end{figure*}

One of the most recent uses of OpenCompass by the Sea-NExT Joint Lab has been to track developments in the sizes of recently released Foundation Models vs. their scores on the OpenCompass benchmark. Figure~\ref{fig:OpenC} illustrates the performance of 24 LLMs released between October 2023 and June 2024, with time represented on the x-axis~\cite{yao2024minicpm}. This visualization is one of the outputs from the OpenCompass evaluation. The y-axis of Figure~\ref{fig:OpenC} is a non-linear depiction of model sizes with the topmost models having unknown sizes. The colour of the dot for each model indicates its OpenCompass score. The systems listed  include some  that voluntarily requested OpenCompass to compute their scores, partially in order to get better exposure to the research community while others are well-known models to add as landmark references, such as GPT-4o and Gemini.  

In Figure~\ref{fig:OpenC} the dotted red  line and shaded area above that line shows the performance of GPT-4V (version as of 2023.11.16). Since its release in November 2023 we can see at least 2 other models (LLaVA-Next-Yi-34B and InternVL 1.5) equal or exceed its performance, yet they are much smaller in size. Furthermore, for the range of models which are on or below the 10B model size, for the more recent of these models their performances are improving and getting closer to the performance of GPT-4V. 
For example the performance of MiniCPM-Llama3-V2.5 with only about 8B parameters in size and released  in May 2024 and shown as the star in the graph, is very close to GPT-4V.
While this is comparing model performance against just one landmark model and there are many capabilities not fully covered by OpenCompass such as multimodal code, complex math capabilities and multi-image understanding capabilities the trend is clearly evident.  Finally, in Figure~\ref{fig:OpenC} the blue dotted line with the legend entry ``end side computation" that rises slowly over time shows the computation and storage capacity of standalone devices like smartphones, indicating that these smaller models with their improved OpenCompass scores may shortly be working natively on handheld devices.

\section{Intelligence and the Human Brain}

The human brain is the most complex system on the planet and has held a fascination for scientists for centuries.  Neuroscience, the scientific study of the brain and its behaviour, is a multidisciplinary field which has  been active for the last 100 years.

Early neuroscience used brain probes (EEGs) to measure electrical activity in different parts of the human brain in order to try to ``map it” to determine which areas performed which kinds of task. The first of these investigations was carried out by Hans Berger in July 1924~\cite{doi:10.1080/21646821.2024.2327268}.
Fast forward 100 years and current neuroscience still uses brain imaging (fMRI, PET, CT)
and electrophysiology (EEG, MEG)
as imaging techniques to observe neural behaviours as a subject performs certain tasks. As tasks are performed, neuroscientists use the outputs from the imaging  to build an understanding of the brain's structure, function, physiology and pathology~\cite{noggle2021advances}.
Yet despite 100 years of investigation we still don’t understand, let alone can cure, neurological impairments like motor neuron disease, Parkinson’s, epilepsy, or dementias, so neuroscience has a long way to go.

In the study of Foundation Models, {\it Mechanistic Interpretability}~\cite{bereska2024mechanistic} as mentioned earlier  which involves understanding a model from observations of its behaviour under different input situations is equivalent to imaging and probing of the (human) brain’s neural activation patterns.
The most recent work which discovered that combinations of active neurons in Claude are monosemantic~\cite{templeton2024scaling} indicates that while this is a first step, {\it Mechanistic Interpretability} of Foundation Models has a very long way to go, and there is almost no guidance on a direction from 100 years of experimental research in neuroscience.

One consolation for us here is that there are major differences between the human brain and a Foundation Model.
The human brain has about 86 Billion neurons and up to 100 trillion synapses, which we could equate to parameters while the largest LLM has 0.3\% of such connections.
Brain synapses are far more complex than the fixed links with weights which we have in Foundation Models, involving numerous biochemical processes. 
Synapses continuously adapt through neural plasticity, while Foundation Model links are fixed post-training, except for when fine tuning occurs.
Brain synapses are energy-efficient compared to Foundation Models and neurons and synapses die and new ones generate as we age.
Finally, the brain has homeostatic regulation in that it maintains overall balance of excitation and activation levels (except during seizures) for energy management and it regulates circadian rhythm whereas Foundation Models, as we are just discovering, have irregular bursts of activation~\cite{templeton2024scaling}.  So with all these differences perhaps we should not turn to neuroscience research for guidance in trying to understand Foundation Models.

\section{Conclusions}

This paper has examined the rapid evolution of Foundation Models (FMs) and their growing capabilities in reasoning and knowledge representation. We have identified several key trends including:
\begin{enumerate}
\item FMs are demonstrating increasingly sophisticated reasoning abilities, often surpassing earlier models in tasks requiring common-sense understanding and logical inference.

\item The phenomenon of grokking, where models suddenly exhibit near-perfect performance after extended training, suggests that FMs may be developing deeper, more abstract representations of knowledge.

\item Contrary to initial assumptions, smaller models with efficient training techniques are showing competitive performance against larger counterparts, indicating that model size alone does not determine capability.

\item  The discovery of monosemantic neuron combinations in FMs points to the emergence of structured knowledge representation within these models, analogous to pattern recognition in biological neural networks.

\item Current benchmarking methods, while diverse, still struggle to comprehensively evaluate the full spectrum of FM capabilities, particularly in assessing overall answer quality and human-like reasoning.

\item Finally, despite some similarities, fundamental differences between the structure of FMs and the human warn us  against making direct comparisons or expecting neuroscience to provide immediate insights into FM function.
\end{enumerate}

These trends collectively suggest that FMs are progressing towards more efficient, interpretable, and potentially more ``intelligent" systems. However, our understanding of their internal mechanisms remains limited, much like our ongoing efforts to comprehend the human brain.

As FMs continue to advance, future research should focus on developing more comprehensive evaluation methods and benchmarks, improving model interpretability, and exploring the ethical implications of increasingly capable AI systems. 

\section*{Acknowledgment}

This paper is the result of discussions with  Prof. Tat-Seng Chua of the Sea-NExT Joint Lab, National University of Singapore, Singapore. 

\bibliographystyle{plain} 
\bibliography{refs}

\begin{thebibliography}{10}

\bibitem{abdel2014convolutional}
Ossama Abdel-Hamid, Abdel-rahman Mohamed, Hui Jiang, Li~Deng, Gerald Penn, and Dong Yu.
\newblock Convolutional neural networks for speech recognition.
\newblock {\em IEEE/ACM Transactions on audio, speech, and language processing}, 22(10):1533--1545, 2014.

\bibitem{bereska2024mechanistic}
Leonard Bereska and Efstratios Gavves.
\newblock {Mechanistic Interpretability for AI Safety--A Review}.
\newblock {\em arXiv preprint arXiv:2404.14082}, 2024.

\bibitem{bisk2020piqa}
Yonatan Bisk, Rowan Zellers, Jianfeng Gao, Yejin Choi, et~al.
\newblock {PIQA: Reasoning about physical commonsense in natural language}.
\newblock In {\em Proceedings of the AAAI conference on artificial intelligence}, volume~34, pages 7432--7439, 2020.

\bibitem{carolan2024reviewmultimodallargelanguage}
Kilian Carolan, Laura Fennelly, and Alan~F. Smeaton.
\newblock {A Review of Multi-Modal Large Language and Vision Models}.
\newblock {\em arXiv preprint arXiv:2404.01322}, 2024.

\bibitem{chiangchatbot}
Wei-Lin Chiang, Lianmin Zheng, Ying Sheng, Anastasios~Nikolas Angelopoulos, Tianle Li, Dacheng Li, Banghua Zhu, Hao Zhang, Michael Jordan, Joseph~E Gonzalez, et~al.
\newblock {Chatbot Arena: An Open Platform for Evaluating LLMs by Human Preference}.
\newblock In {\em Forty-first International Conference on Machine Learning}, July 2024.

\bibitem{clark2019boolq}
Christopher Clark, Kenton Lee, Ming-Wei Chang, Tom Kwiatkowski, Michael Collins, and Kristina Toutanova.
\newblock {BoolQ: Exploring the surprising difficulty of natural yes/no questions}.
\newblock {\em arXiv preprint arXiv:1905.10044}, 2019.

\bibitem{clark2018think}
Peter Clark, Isaac Cowhey, Oren Etzioni, Tushar Khot, Ashish Sabharwal, Carissa Schoenick, and Oyvind Tafjord.
\newblock Think you have solved question answering? try arc, the {AI}2 reasoning challenge.
\newblock {\em arXiv preprint arXiv:1803.05457}, 2018.

\bibitem{elo1961new}
Arpad~E Elo.
\newblock {New USCF rating system}.
\newblock {\em Chess Life}, 16:160--161, 1961.

\bibitem{gunjal2023detecting}
Anisha Gunjal, Jihan Yin, and Erhan Bas.
\newblock Detecting and preventing hallucinations in large vision language models.
\newblock {\em arXiv preprint arXiv:2308.06394}, 2023.

\bibitem{hendrycks2020measuring}
Dan Hendrycks, Collin Burns, Steven Basart, Andy Zou, Mantas Mazeika, Dawn Song, and Jacob Steinhardt.
\newblock Measuring massive multitask language understanding.
\newblock {\em arXiv preprint arXiv:2009.03300}, 2020.

\bibitem{jiang2023mistral}
Albert~Q Jiang, Alexandre Sablayrolles, Arthur Mensch, Chris Bamford, Devendra~Singh Chaplot, Diego de~las Casas, Florian Bressand, Gianna Lengyel, Guillaume Lample, Lucile Saulnier, et~al.
\newblock {Mistral 7B}.
\newblock {\em arXiv preprint arXiv:2310.06825}, 2023.

\bibitem{karystinos2000overfitting}
George~N Karystinos and Dimitrios~A Pados.
\newblock On overfitting, generalization, and randomly expanded training sets.
\newblock {\em IEEE Transactions on Neural Networks}, 11(5):1050--1057, 2000.

\bibitem{lin2022truthfulqa}
Stephanie Lin, Jacob Hilton, and Owain Evans.
\newblock {TruthfulQA: Measuring How Models Mimic Human Falsehoods}.
\newblock In {\em Proceedings of the 60th Annual Meeting of the Association for Computational Linguistics (Volume 1: Long Papers)}, pages 3214--3252, 2022.

\bibitem{liu2022towards}
Ziming Liu, Ouail Kitouni, Niklas~S Nolte, Eric Michaud, Max Tegmark, and Mike Williams.
\newblock Towards understanding grokking: An effective theory of representation learning.
\newblock {\em Advances in Neural Information Processing Systems}, 35:34651--34663, 2022.

\bibitem{lu2007improving}
Yajuan L{\"u}, Jin Huang, and Qun Liu.
\newblock Improving statistical machine translation performance by training data selection and optimization.
\newblock In {\em Proceedings of the 2007 Joint Conference on Empirical Methods in Natural Language Processing and Computational Natural Language Learning (EMNLP-CoNLL)}, pages 343--350, 2007.

\bibitem{mihaylov2018can}
Todor Mihaylov, Peter Clark, Tushar Khot, and Ashish Sabharwal.
\newblock {Can a Suit of Armor Conduct Electricity? A New Dataset for Open Book Question Answering}.
\newblock In {\em Proceedings of the 2018 Conference on Empirical Methods in Natural Language Processing}, pages 2381--2391, 2018.

\bibitem{noggle2021advances}
Chad~A Noggle and Andrew~S Davis.
\newblock Advances in neuroimaging.
\newblock {\em Understanding the Biological Basis of Behavior: Developing Evidence-Based Interventions for Clinical, Counseling and School Psychologists}, pages 107--137, 2021.

\bibitem{doi:10.1080/21646821.2024.2327268}
Pooja Ojha.
\newblock {Berger and the Breakthrough: A Centennial Celebration of the Human Electroencephalogram}.
\newblock {\em The Neurodiagnostic Journal}, 64(2):69--74, 2024.
\newblock PMID: 38772013.

\bibitem{OpenCompass}
OpenCompass.
\newblock {Large Model Evaluation System: Opencompass}.
\newblock \url{https://opencompass.org.cn/home}.
\newblock Accessed: 2024-06-28.

\bibitem{CBA}
Large Model Systems Organization~(LMSYS Org).
\newblock {LMSYS Chatbot Arena Leaderboard}.
\newblock \url{https://chat.lmsys.org/?leaderboard}, Last updated: 2024-06-26.
\newblock Accessed: 2024-06-28.

\bibitem{seymour2011history}
Tom Seymour, Dean Frantsvog, Satheesh Kumar, et~al.
\newblock History of search engines.
\newblock {\em International Journal of Management \& Information Systems (IJMIS)}, 15(4):47--58, 2011.

\bibitem{templeton2024scaling}
Adly Templeton, Tom Conerly, Jonathan Marcus, Jack Lindsey, Trenton Bricken, Brian Chen, Adam Pearce, Craig Citro, Emmanuel Ameisen, Andy Jones, Hoagy Cunningham, Nicholas~L Turner, Callum McDougall, Monte MacDiarmid, C.~Daniel Freeman, Theodore~R. Sumers, Edward Rees, Joshua Batson, Adam Jermyn, Shan Carter, Chris Olah, and Tom Henighan.
\newblock Scaling monosemanticity: Extracting interpretable features from claude 3 sonnet.
\newblock {\em Transformer Circuits Thread}, 2024.

\bibitem{vaswani2017attention}
Ashish Vaswani, Noam Shazeer, Niki Parmar, Jakob Uszkoreit, Llion Jones, Aidan~N Gomez, {\L}ukasz Kaiser, and Illia Polosukhin.
\newblock Attention is all you need.
\newblock {\em Advances in neural information processing systems}, 30, 2017.

\bibitem{xiao2023smoothquant}
Guangxuan Xiao, Ji~Lin, Mickael Seznec, Hao Wu, Julien Demouth, and Song Han.
\newblock Smoothquant: Accurate and efficient post-training quantization for large language models.
\newblock In {\em International Conference on Machine Learning}, pages 38087--38099. PMLR, 2023.

\bibitem{10.1145/3649506}
Jingfeng Yang, Hongye Jin, Ruixiang Tang, Xiaotian Han, Qizhang Feng, Haoming Jiang, Shaochen Zhong, Bing Yin, and Xia Hu.
\newblock {Harnessing the Power of LLMs in Practice: A Survey on ChatGPT and Beyond}.
\newblock {\em ACM Trans. Knowl. Discov. Data}, 18(6), apr 2024.

\bibitem{yao2024minicpm}
Yuan Yao, Tianyu Yu, Ao~Zhang, Chongyi Wang, Junbo Cui, Hongji Zhu, Tianchi Cai, Haoyu Li, Weilin Zhao, Zhihui He, et~al.
\newblock {MiniCPM-V: A GPT-4V Level MLLM on Your Phone}.
\newblock {\em arXiv preprint arXiv:2408.01800}, 2024.

\bibitem{zellers2019hellaswag}
Rowan Zellers, Ari Holtzman, Yonatan Bisk, Ali Farhadi, and Yejin Choi.
\newblock {Hellaswag: Can a machine really finish your sentence?}
\newblock {\em arXiv preprint arXiv:1905.07830}, 2019.

\bibitem{zhang2023counterfactual}
Chiyuan Zhang, Daphne Ippolito, Katherine Lee, Matthew Jagielski, Florian Tram{\`e}r, and Nicholas Carlini.
\newblock Counterfactual memorization in neural language models.
\newblock {\em Advances in Neural Information Processing Systems}, 36:39321--39362, 2023.

\bibitem{Zhao2023ASO}
Wayne~Xin Zhao, Kun Zhou, Junyi Li, Tianyi Tang, Xiaolei Wang, Yupeng Hou, Yingqian Min, Beichen Zhang, Junjie Zhang, Zican Dong, Yifan Du, Chen Yang, Yushuo Chen, Z.~Chen, Jinhao Jiang, Ruiyang Ren, Yifan Li, Xinyu Tang, Zikang Liu, Peiyu Liu, Jianyun Nie, and Ji~rong Wen.
\newblock {A Survey of Large Language Models}.
\newblock {\em ArXiv}, abs/2303.18223, 2023.

\end{thebibliography}

\end{document}